\pdfoutput=1 

\documentclass[11pt,letterpaper]{article}
\usepackage{emnlp2016}		
\usepackage{times}
\usepackage{latexsym}
\usepackage{amsmath}
\usepackage{fancyvrb, fancyhdr, theorem, latexsym, color, longtable}
\usepackage{multirow}
\usepackage{url}
\usepackage{bm}
\usepackage{amssymb}
\usepackage{fixltx2e}
\usepackage{tabularx}
\usepackage{hyperref}
\usepackage{graphicx}
\usepackage{color}

\emnlpfinalcopy

\def\emnlppaperid{392}

\title{Creating Causal Embeddings for Question Answering \\with Minimal Supervision
}

\author{Rebecca Sharp, Mihai Surdeanu, Peter Jansen, Peter Clark, \and Michael Hammond \\ University of Arizona, Allen Institute for Artificial Intelligence \\ \{bsharp, msurdeanu, pajansen, hammond\}@email.arizona.edu, PeterC@allenai.org}

\begin{document}
\vspace{-5mm}
\maketitle

\begin{abstract}
A common model for question answering (QA) is that a good answer is one that is closely related to the question, where relatedness is often determined using general-purpose lexical models such as word embeddings. 
We argue that a better approach is to look for answers that are related to the question in a {\em relevant way}, according to the information need of the question,
which may be determined through task-specific embeddings. 
With causality as a use case, we implement this insight in three steps. First, we generate causal embeddings cost-effectively by bootstrapping cause-effect pairs extracted from free text using a small set of seed patterns. Second, we train dedicated embeddings over this data, by using task-specific contexts, i.e., the context of a cause is its effect. Finally, we extend a state-of-the-art reranking approach for QA to incorporate these causal embeddings. We evaluate the causal embedding models both \emph{directly} with a casual implication task,
 and \emph{indirectly}, in a downstream causal QA task using data from Yahoo! Answers. We show that explicitly modeling causality improves performance in both tasks. In the QA task our best model achieves 37.3\% P@1, significantly outperforming a strong baseline by 7.7\% (relative). 

\end{abstract}

\section{Introduction}
\label{sec:introduction}

Question answering (QA), i.e., finding short answers to natural language questions, is one of the most important but challenging 
tasks on the road towards natural language understanding~\cite{Etzioni:11}. 
A common approach for QA is to prefer answers that are closely related to the question, where relatedness is often determined using lexical semantic models such as word embeddings~\cite{yih13,jansen14,fried2015higher}. 
While appealing for its robustness to natural language variation, this one-size-fits-all approach does not take into account the wide range of distinct question types that can appear in any given question set, and that are best addressed individually~\cite{chu2004ibm,ferrucci2010building,clark2013study}.  

Given the variety of question types, we suggest that a better approach is to look for answers 
that are related to the question \emph{through the appropriate relation}, e.g., a causal question should have a cause-effect relation with its answer.
If we adopt this view, and continue to work with embeddings as a mechanism for assessing relationship,
this raises a key question: how do we train and use task-specific embeddings cost-effectively? 
Using causality as a use case, we answer this question with a framework for producing causal word embeddings with minimal supervision, and a demonstration that such task-specific embeddings significantly benefit causal QA. 

In particular, the contributions of this work are:

{\flushleft {\bf (1)}} 
A methodology for generating causal embeddings cost-effectively by bootstrapping cause-effect pairs extracted from free text using a small set of seed patterns, e.g., {\em X causes Y}. 
We then train dedicated embedding (as well as two other distributional similarity) models over this data. Levy and Goldberg~\shortcite{levy2014dependency} have modified the algorithm of Mikolov et al.~\shortcite{mikolov2013distributed} to use an arbitrary, rather than linear, context. Here we make this context task-specific, i.e., the context of a cause is its effect.
Further, to mitigate sparsity and noise, our models are bidirectional, and noise aware (by incorporating the likelihood of noise in the training process). 

{\flushleft {\bf (2)}} The insight that QA benefits from task-specific embeddings. 
We implement a QA system that uses the above causal embeddings to answer questions and demonstrate that they significantly improve performance over a strong baseline. Further, we show that causal embeddings encode complementary information to vanilla embeddings, even when trained from the same knowledge resources. 

{\flushleft {\bf (3)}} An analysis of direct vs. indirect evaluations for task-specific word embeddings. 
We evaluate our causal models both  {\em directly}, in terms of measuring their capacity to rank causally-related word pairs over word pairs of other relations, as well as {\em indirectly} in the downstream causal QA task. 
In both tasks, our analysis indicates that including causal models significantly improves performance. 
However, from the direct evaluation, it is difficult to estimate which models will perform best in real-world tasks. Our analysis re-enforces recent observations about the limitations of word similarity evaluations~\cite{faruqui2016problems}: we show that they have limited coverage and may align poorly with real-world tasks.

\section{Related Work}
\label{sec:related work}

Addressing the need for specialized solving methods in QA, 
Oh et. al~\shortcite{oh2013question} incorporate a dedicated causal component into their system, and note that it improves the overall performance.  However, their model is limited by the need for lexical overlap between a causal construction found in their knowledge base and the question itself.  Here, we develop a causal QA component that exploits specialized word embeddings to gain robustness to lexical variation.  

There has been a vast body of work which demonstrates that word embeddings derived from distributional similarity are useful in many tasks, including question answering -- see \emph{inter alia}~\mbox{\cite{fried2015higher,yih13}}.  However, Levy and Goldberg~\shortcite{levy2015supervised} note that there are limitations on the type of semantic knowledge which is encoded in these general-purpose similarity embeddings. 
Therefore, here we build customized task-specific embeddings for causal QA.

Customized embeddings have been created for a variety of tasks, including semantic role labeling~\cite{fitzgerald2015semantic,woodsenddistributed}, and binary relation extraction ~\mbox{\cite{riedel2013relation}.}
Similar to Riedel et al., we train embeddings customized for specific relations, but we bootstrap training data using minimal supervision (i.e., a small set of patterns) rather than relying on distant supervision and large existing knowledge bases.  Additionally, while Riedel et al. represent all relations in a general embedding space, here we train a dedicated embedding space for just the causal relations. 

In QA, embeddings have been customized to have question words that are close to either their answer words~\cite{bordes2014question}, or to structured knowledge base entries~\cite{yang2014joint}.  While these methods are useful for QA, they do not distinguish between different types of questions, and as such their embeddings are not specific to a given question type.

Additionally, embeddings have been customized to distinguish functional similarity from relatedness ~\cite{levy2014dependency,kielaspecializing}.
In particular, Levy and Goldberg train their embeddings by replacing the standard linear context of the target word with context derived from the syntactic dependency graph of the sentence.  
In this work, we make use of this extension to arbitrary context in order to train our embeddings with contexts derived from binary causal relations.  We extract cause-effect text pairs such that the cause text becomes the \emph{target} text and the effect text serves as the \emph{context}. 

Recently, Faruqui et al.\shortcite{faruqui2016problems} discussed issues surrounding the evaluation of similarity word embeddings, including the lack of correlation between their performance on word-similarity tasks and ``downstream'' or real-world tasks like QA, text classification, etc.  As they advocate, in addition to a direct evaluation of our causal embeddings, we also evaluate them independently in a downstream QA task.  We provide the same comparison for two alternative approaches (an alignment model and a convolutional neural network model), confirming that the direct evaluation performance can be misleading without the task-specific, downstream evaluation. 

With respect to extracting causal relations from text, Girju et al.~\shortcite{girju2002text} use modified Hearst patterns~\cite{hearst1992automatic} to extract a large number of potential cause-effect tuples, where both causes and effects must be nouns.
However, Cole et al.~\shortcite{cole2005lightweight} show that these nominal-based causal relations account for a relatively small percentage of all causal relations, 
and for this reason, \cite{yang2014multi} allow for more elaborate argument structures in their causal extraction by identifying verbs, and then following the syntactic subtree of the verbal arguments to construct their candidate causes and effects. 
Additionally, Do et al.~\shortcite{do2011minimally} observe that nouns as well as verbs can signal causality.  
We follow these intuitions in developing our causal patterns by using both nouns and verbs to signal potential participants in causal relations, and then allowing for the entire dominated structures to serve as the cause and/or effect arguments.

\section{Approach}
\label{sec:approach}

Our focus is on reranking answers to causal questions using using task-specific distributional similarity methods.
Our approach operates in three steps:

{\flushleft (1)} We start by bootstrapping a large number of cause-effect pairs from free text using a small number of syntactic and surface patterns (Section \ref{sec:causalextraction}).

{\flushleft (2)} We then use these bootstrapped pairs to build several task-specific embedding (and other distributional similarity) models (Section \ref{sec:models}). We evaluate these models directly on a causal-relation identification task (Section \ref{sec:directeval}).  

{\flushleft (3)} Finally, we incorporate these models into a reranking framework for causal QA and demonstrate that the resulting approach performs better than the reranker without these task-specific models, even if trained on the same data (Section ~\ref{sec:indirecteval}).

\section{Extracting Cause-Effect Tuples}
\label{sec:causalextraction}

Because the success of embedding models depends on large training datasets \cite{sharp2015spinning}, and such datasets do not exist for open-domain causality, we opted to bootstrap a large number of cause-effect pairs from a small set of patterns.
We wrote these patterns using Odin~\cite{valenzuela2016runes}, a rule-based information extraction framework which has the distinct advantage of 
being able to operate over multiple representations of content (i.e., surface and syntax).
For this work, we make use of rules that operate over both surface sequences as well as dependency syntax in the grammars introduced in steps (2) and (3) below.

Odin operates as a cascade, 
allowing us to implement a two-stage approach.
First, we identify potential participants in causal relations, i.e., the potential causes and effects, which we term {\bf causal mentions (CM)}. A second grammar then identifies actual causal relations that take these CMs as arguments.

We consider both noun phrases (NP) as well as entire 
clauses to be potential CMs, since causal patterns form around participants that are syntactically more complex than flat NPs.  
For example, in the sentence \emph{The collapse of the housing bubble caused stock prices to fall}, both the cause ({\em the collapse of the housing bubble}) and effect ({\em stock prices to fall}) are more complicated nested structures.  Reducing these arguments to non-recursive NPs (e.g., {\em The collapse} and {\em stock prices}) is clearly insufficient to capture the relevant context.

Formally, we extract our causal relations using the following algorithm:
{\flushleft \textbf{(1) Pre-processing:}} Much of the text we use to extract causal relation tuples comes from the Annotated Gigaword \cite{napoles2012annotated}.  This text is already fully annotated and no further processing is necessary.  We additionally use text from the Simple English Wikipedia\footnote{{\scriptsize \url{https://simple.wikipedia.org/wiki/Main_Page}}.  The Simple English version was preferred over the full version due to its simpler sentence structures, which make extracting cause-effect tuples more straightforward.}, which we processed using the Stanford CoreNLP toolkit~\cite{Manning:14} and the dependency parser of Chen and Manning~\shortcite{chen14}.

{\flushleft \textbf{(2) CM identification:}} \label{step:cm} We extract causal mentions (which are able to serve as arguments in our causal patterns) using a set of rules  designed to be robust to the variety that exists in natural language. 
Namely, to find CMs that are noun phrases, we first find words that are tagged as nouns, then follow outgoing dependency links for modifiers and attached prepositional phrases\footnote{The outgoing dependency links from the nouns which we followed were: \texttt{nn, amod, advmod, ccmod, dobj, prep\_of, prep\_with, prep\_for, prep\_into, prep\_on, prep\_to}, and \texttt{prep\_in}.}, to a maximum depth of two links.  To find CMs that are clauses, we first find words that are tagged as verbs (excluding verbs which themselves were considered to signal causation\footnote{The verbs we excluded were: \emph{cause, result, lead, create}.}), then again follow outgoing dependency links for modifiers and arguments.  We used a total of four rules to label CMs.

{\flushleft \textbf{(3) Causal tuple extraction:}} \label{step:causalext} After CMs are identified, a grammar scans the text for causal relations that have CMs as arguments.  Different patterns have varying probabilities of signaling causation~\cite{khoo1998automatic}.  To minimize the noise in the extracted pairs, we restrict ourselves to a set of 13 rules designed to find unambiguously causal patterns, such as {\em CAUSE led to EFFECT}, where {\em CAUSE} and {\em EFFECT} are CMs.
The rules operate by looking for a \emph{trigger} phrase, e.g., {\em led}, and then following the dependency paths to and/or from the trigger phrase to see if all required CM arguments exist.

\begin{table}[t!]
\begin{center}
\begin{footnotesize}
\begin{tabular}{lc}
\hline
Corpus		&	Extracted Tuples		 \\
\hline
Annotated Gigaword	& 798,808 	\\
Simple English Wikipedia		& 16,425 	\\
\hline
Total		& 815,233 	\\
\end{tabular}
\end{footnotesize}
\caption{{\footnotesize Number of causal tuples extracted from each corpus.}} 
\label{tab:causalstats}
\vspace{-4mm}
\end{center}
\end{table}

Applying this causal grammar over Gigaword and Simple English Wikipedia produced 815,233 causal tuples, as summarized in Table~\ref{tab:causalstats}. As bootstrapping methods are typically noisy, we manually evaluated the quality of approximately 250 of these pairs selected at random.  Of the tuples evaluated, approximately 44\% contained some amount of noise. For example, from the sentence \emph{Except for Springer's show, which still relies heavily on confrontational topics that lead to fistfights virtually every day...}, while ideally we would only extract (\emph{confrontational topics $\rightarrow$ fistfights}), instead we extract the tuple (\emph{show which still relies heavily on confrontational topics} $\rightarrow$ \emph{fistfights virtually every day}), which contains a large amount of noise: \emph{show, relies, heavily}, etc.
This finding prompted our noise-aware model described at the end of Section~\ref{sec:models}.

\section{Models}
\label{sec:models}

We use the extracted causal tuples to train three distinct distributional similarity models that explicitly capture causality. 

{\flushleft \textbf{Causal Embedding Model (cEmbed):}}
The first distributional similarity model we use is based on the skip-gram word-embedding algorithm of Mikolov et al.~\shortcite{mikolov2013distributed}, which has been shown to improve a variety of language processing tasks 
including QA~\cite{yih13,fried2015higher}.  In particular, we use the variant implemented by Levy and Goldberg~\shortcite{levy2014dependency} which modifies the original algorithm to use an arbitrary, rather than linear, context. 
Our novel contribution is to make this context task-specific: intuitively, the context of a cause is its effect. Further, these contexts are generated from tuples that are themselves bootstrapped, which minimizes the amount of supervision necessary.

The Levy and Goldberg model trains using single-word pairs, while our CMs could be composed of multiple words.  
For this reason, we decompose each cause--effect tuple, $(CM_c,CM_e)$, such that each word $w_c \in CM_c$ is paired with each word $w_e \in CM_e$. 

After filtering the extracted cause-effect tuples for stop words and retaining only nouns, verbs, and adjectives, we generated over 3.6M $(w_c, w_e)$ word-pairs\footnote{For all models proposed in this section we used lemmas rather than words.} from the approximately 800K causal tuples.

The model learns two embedding vectors for each word, one for when the word serves as a target word and another for when the word serves as a context word.  Here, since the relation of interest is inherently directional, both sets of embeddings are meaningful, and so we make use of both -- the target vectors encode the effects of given causes, whereas the context vectors capture the causes of the corresponding effects.

{\flushleft \textbf{Causal Alignment Model (cAlign):}}
Monolingual alignment (or translation) models have been shown to be successful in QA \cite{Berger:00,Echihabi:03,Soricut:06,Riezler:etal:2007,Surdeanu:11,yao2013}, and recent work has shown that they can be successfully trained with less data than embedding models~\cite{sharp2015spinning}. 

To verify these observations in our context, we train an alignment model that ``translates'' causes (i.e., the ``source language'') into effects (i.e., the ``destination language''), using our cause--effect tuples. 
This is done using IBM Model 1~\cite{Brown:93} and GIZA++~\cite{och03}.

\begin{figure}[t!]
\begin{center}
\includegraphics[width=60mm]{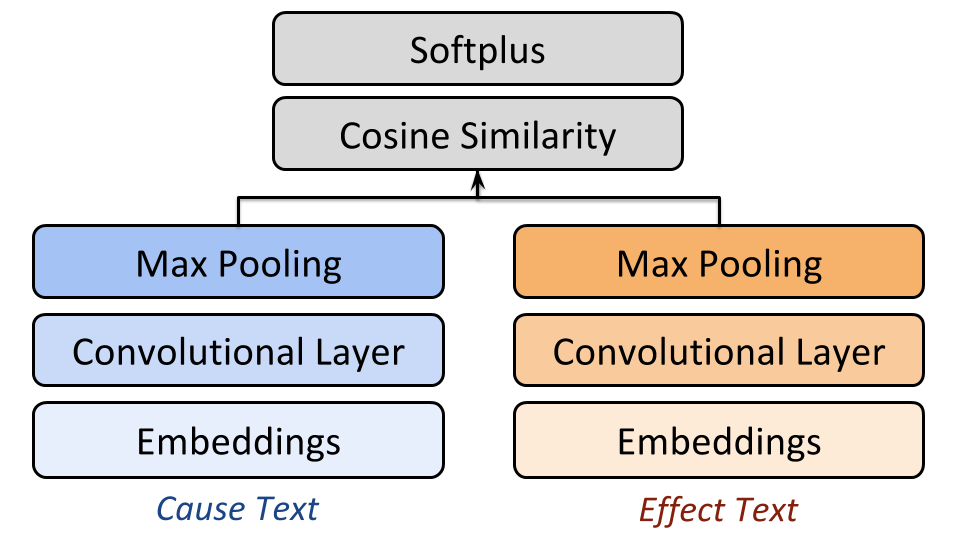}
\vspace{-2mm}
\caption{{\footnotesize Architecture of the causal convolutional network. }}
\vspace{-6mm}
\label{fig:cnn}
\end{center}
\end{figure}

{\flushleft \textbf{Causal Convolutional Neural Network Model (cCNN):}}
Each of the previous models have at their root a bag-of-words representation, which is a simplification of the causality task. To address this potential limitation, we additionally trained a convolutional neural network (CNN) which operates over variable-length texts, and maintains distinct embeddings for causes and effects.  The architecture of this approach is shown in Figure~\ref{fig:cnn}, and consists of two sub-networks (one for cause text and one for effect text), each of which begins by converting the corresponding text 
into 50-dimensional embeddings.  These are then fed to a convolutional layer,\footnote{The convolutional layer contained 100 filters, had a filter length of 2 (i.e., capturing bigram information), and an inner ReLU activation.} which is followed by a max-pooling layer of equal length.
Then, these top sub-network layers, which can be thought of as a type of phrasal embedding, are merged by taking their cosine similarity.  Finally, this cosine similarity is normalized by feeding it into a dense layer with a single node which has a softplus activation.  
In designing our CNN, we attempted to minimize architectural and hyperparameter tuning by taking inspiration from Iyyer et al.~\shortcite{iyyer2015deep}, preferring simpler architectures.
We train the network using a binary cross entropy objective function and the Adam optimizer~\cite{kingma2014adam}, using the Keras library~\cite{chollet2015keras} operating over Theano~\cite{2016arXiv160502688short}, a popular deep-learning framework.\footnote{We also experimented with an equivalent architecture where the sub-networks are implemented using long short-term memory (LSTM) networks~\cite{hochreiter1997long}, and found that they consistently under-perform this CNN architecture. Our conjecture is that CNNs perform better because LSTMs are more sensitive to overall word order than CNNs, which capture only local contexts, and we have relatively little training data, which prevents the LSTMs from generalizing well.}

{\flushleft \textbf{Noise-aware Causal Embedding Model (cEmbedNoise):}} 
We designed a variant of our cEmbed approach to address the potential impact of the noise introduced by our bootstrapping method.
While training, we weigh the causal tuples by the likelihood that they are truly causal, which we approximate with pointwise mutual information (PMI).
For this, we first score the tuples by their causal PMI and then scale these scores by the overall frequency of the tuple~\cite{riloff1996automatically}, to account for the PMI bias toward low-frequency items.  That is, the score $S$ of a tuple, $t$, is computed as: 

\vspace{-2mm}
\begin{small}
\begin{equation}
S(t) = \log \frac{p(t|causal)}{p(t)} * \log (freq(t))
\end{equation} 
\end{small}
\vspace{-2mm}

We then discretize these scores into five quantiles, ascribing a linearly decreasing weight during training to datums in lower scoring quantiles.

\section{Direct Evaluation: Ranking Word Pairs}

\begin{figure*}[th!]
\begin{center}
\includegraphics[width=0.50\textwidth]{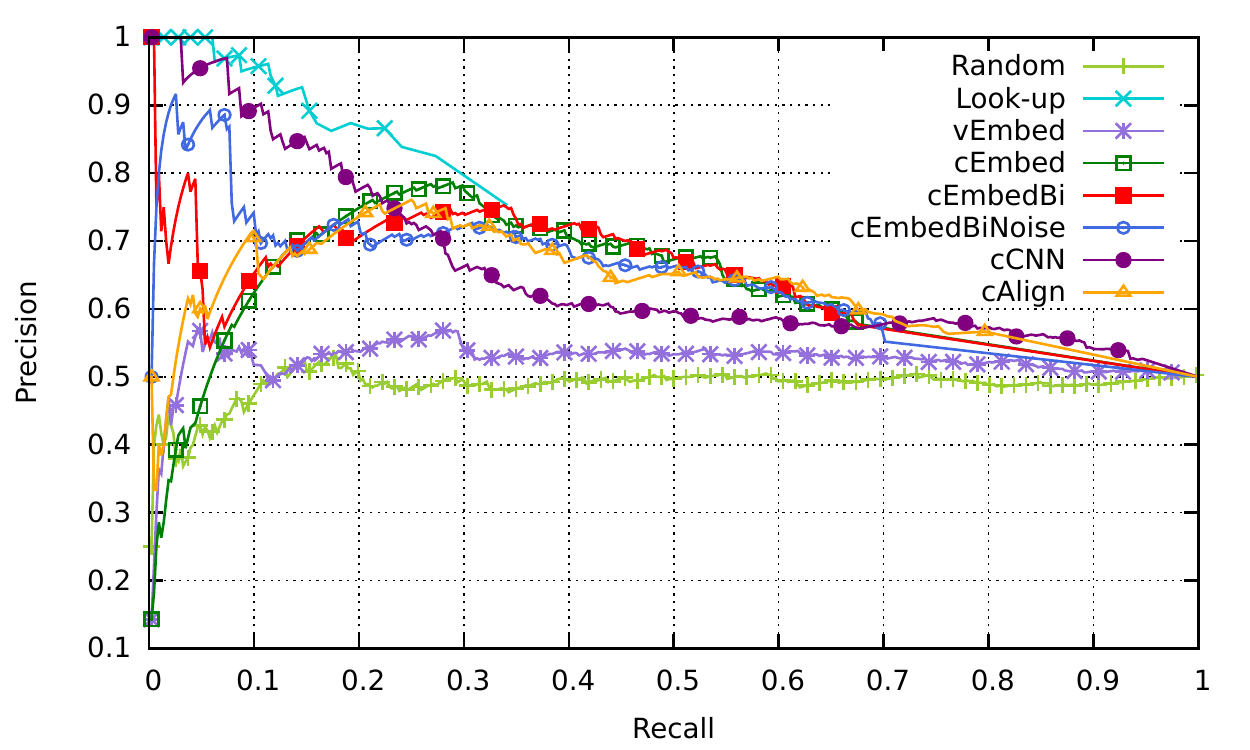} 
\vspace{-3mm}
\caption{{\footnotesize Precision-recall curve showing the ability of each model to rank causal pairs above non-causal pairs. For clarity, we do not plot cEmbedNoise, which performs worse than cEmbedBiNoise. The Look-up model has no data points beyond the 35\% recall point.}}
\vspace{-4mm}
\label{fig:rpcurve_all}
\end{center}
\end{figure*}

\label{sec:directeval}

We begin the assessment of our models with a {\em direct} evaluation to determine whether or not the proposed approaches capture causality better than general-purpose word embeddings and whether their robustness improves upon a simple database look-up.
For this evaluation, we follow the protocol of Levy and Goldberg~\shortcite{levy2014dependency}.  
In particular, we create a collection of word pairs, half of which are causally related, with the other half consisting of other relations. 
These pairs are then ranked by our models and several baselines, with the goal of ranking the causal pairs above the others. 
The embedding models rank the pairs using the cosine similarity between the target vector for the causal word and the context vector of the effect word.  The alignment model ranks pairs using the probability $P(\text{Effect}|\text{Cause})$ given by IBM Model 1, and the CNN ranks pairs by the value of the output returned by the network.

\subsection{Data}
In order to avoid bias towards our extraction methods, we evaluate our models on an external set of word pairs drawn from the SemEval 2010 Task 8 \cite{hendrickx2009semeval}, originally a multi-way classification of semantic relations between nominals.  We used a total of 1730 nominal pairs, 865 of which were from the Cause-Effect relation (e.g., (\emph{dancing $\rightarrow$ happiness})) and an equal number which were randomly selected from the other eight relations (e.g., (\emph{juice $\rightarrow$ grapefruit}), from the Entity-Origin relation).  This set was then randomly divided into equally-sized development and test partitions.

\subsection{Baselines}
We compared our distributional similarity models against three baselines:

{\flushleft \textbf{Vanilla Embeddings Model (vEmbed):}} a standard \texttt{word2vec} model trained with the skip-gram algorithm and a sliding window of 5, using the original texts from which our causal pairs were extracted.\footnote{All embedding models analyzed here, including this baseline and our causal variants, produced embedding vectors of 200 dimensions.} As with the cEmbed model, SemEval pairs were ranked using the cosine similarity between the vector representations of their arguments.
\vspace{-1mm}

{\flushleft \textbf{Look-up Baseline:}} a given SemEval pair was ranked by the number of times it appeared in our extracted cause-effect tuples. 
\vspace{-1mm}

{\flushleft \textbf{Random:}} pairs were randomly shuffled.
\vspace{-1mm}

\subsection{Results}

Figure \ref{fig:rpcurve_all} shows the precision-recall (PR) curve for each of the models and baselines. 
As expected, the causal models are better able to rank causal pairs than the vanilla embedding baseline (vEmbed), which, in turn, outperforms the random baseline.  Our look-up baseline, which ranks pairs by their frequency in our causal database, shows a high precision for this task, but has coverage for only 35\% of the causal SemEval pairs.

Some models perform better on the low-recall portion of the curve (e.g., the look-up baseline and cCNN), while the embedding and alignment models have a higher and more consistent performance across the PR curve. We hypothesize that models that better \emph{balance} precision and recall will perform better in a real-world QA task, which may need to access a given causal relation through a variety of lexical patterns or variations. We empirically validate this observation in Section~\ref{sec:indirecteval}.

The PR curve for the causal embeddings shows an atypical dip at low-recall.  To examine this, we analyzed its top-ranked 15\% of SemEval pairs, and found that incorrectly ranked pairs were not found in the database of causal tuples.  Instead, these incorrect rankings were largely driven by low frequency words whose embeddings could not be robustly estimated due to lack of direct evidence.  
Because this sparsity is partially driven by directionality, 
we implemented a bidirectional embedding model (cEmbedBi) that (a) trains a second embedding model by reversing the input (effects as targets, causes as contexts), and (b) ranks pairs by the \textit{average} of the scores returned by these two unidirectional causal embedding models. 
Specifically, the final bidirectional score of the pair, $(e_1, e_2)$, where $e_1$ is the candidate cause and $e_2$ is the candidate effect, is:
\begin{equation}
s_{bi}(e_1, e_2) = \tfrac{1}{2}(s_{c{\tiny \rightarrow}e}(e_1, e_2) + s_{e \rightarrow c}(e_2, e_1))
\end{equation}
where $s_{c \rightarrow e}$ is the score given by the original causal embeddings, i.e., from cause to effect, and $s_{e \rightarrow c}$ is the score given by the reversed-input causal embeddings, i.e., from effect to cause.

As Figure~\ref{fig:rpcurve_all} shows, the bidirectional embedding variants consistently outperform their unidirectional counterparts. 
All in all, the best overall model is cEmbedBiNoise, which is both bidirectional and incorporates the noise handling approach from Section~\ref{sec:models}. This model substantially improves performance in the low-recall portion of the curve, while also showing strong performance across the curve.

\vspace{-1mm}
\section{Indirect Evaluation: QA Task}
\vspace{-1mm}
\label{sec:indirecteval}

The main objective of our work is to investigate the impact of a customized causal embedding model for QA. Following our direct evaluation, which solely evaluated the degree to which our models directly encode causality, here we evaluate each of our proposed causal models in terms of their contribution to a downstream real-world QA task.

Our QA system uses a standard reranking approach~\cite{jansen14}.
In this architecture, the candidate answers are initially extracted and ranked using a shallow candidate retrieval (CR) component that uses solely information retrieval techniques, then they are re-ranked using a ``learning to rank'' approach.
In particular, we used SVM rank\footnote{ \url{http://www.cs.cornell.edu/people/tj/svm_light/svm_rank.html}}, a Support Vector Machines classifier adapted for ranking, and re-ranked the candidate answers with a set of features derived from both the initial CR score and the models we have introduced. For our model combinations (see Table \ref{tab:QA}), the feature set includes the CR score and the features from each of the models in the combination.

\subsection{Data}

We evaluate on a set of causal questions extracted from the Yahoo! Answers corpus\footnote{Freely available through Yahoo!'s Webscope
program ({\scriptsize {\tt research-data-requests@yahoo-inc.com}})} with simple surface patterns such as \emph{What causes ...} and ~\emph{What is the result of ...}\footnote{We lightly filtered these with stop words to remove non-causal questions, such as those based on math problems and the results of sporting events. Our dataset will be freely available, conditioned on users having obtained the Webscope license.}.
We extracted a total of 3031 questions, each with at least four candidate answers, and we evaluated performance using five-fold cross-validation, with three folds for training, one for development, and one for testing. 

\subsection{Models and Features}

We evaluate the contribution of the bidirectional and noise-aware causal embedding models (cEmbedBi, and cEmbedBiNoise) as well as the causal alignment model (cAlign) and the causal CNN (cCNN).  These models are compared against three baselines: the vanilla embeddings (vEmbed), the lookup baseline (LU), and additionally a vanilla alignment model (vAlign) which is trained over 65k question-answer pairs from Yahoo! Answers.

The features\footnote{Due to the variety of features used, each feature described here is independently normalized to lie between 0.0 and 1.0.} we use for the various models are:

{ \flushleft \textbf{Embedding model features:}}
For both our vanilla and causal embedding models, we use the same set of features as 
Fried et al.~\shortcite{fried2015higher}: the maximum, minimum, and average pairwise cosine similarity between question and answer words, as well as the overall similarity between the composite question and answer vectors.  
When using the causal embeddings, since the relation is directed, we first determine whether the question text is the cause or the effect\footnote{We do this through the use of simple regular expressions, e.g., "\^~[Ww]hat ([a-z]+ )\{0,3\}cause.+"}, which in turn determines which embeddings to use for the question text and which to use for the candidate answer texts.  For example, in a question such as "\emph{What causes X?}", since \emph{X} is the effect, all cosine similarities would be found using the effect vectors for the question words and the cause vectors for the answer candidate words. 

{\flushleft \textbf{Alignment model features:}} We use the same global alignment probability, $p(Q|A)$ of Surdeanu et al.~\shortcite{Surdeanu:11}. In our causal alignment model, we adapt this to causality as $p(\text{Effect}|\text{Cause})$, and again we first determine the direction of the causal relation implied in the question.  We include the additional undirected alignment features based on Jensen-Shannon distance, proposed more recently by Fried et al.~\shortcite{fried2015higher}, in our vanilla alignment model.  However, due to the directionality inherent in causality, they do not apply to our causal model so there we omit them.

{\flushleft \textbf{Look-up feature:}} For the look-up baseline we count the number of times words from the question and answer appear together in our database of extracted causal pairs, once again after determining the directionality of the questions.  If the total number of matches is over a threshold\footnote{Empirically determined to be 100 matches.  Note that using this threshold performed better than simply using the total number of matches.}, we consider the causal relation to be established and give the candidate answer a score of 1; or a score of 0, otherwise.

\begin{table}[t!]
\begin{center}
\begin{footnotesize}
\begin{tabular}{lll}
\hline
\# & Model & P@1 \\ 
\hline
& Baselines & \\
\hline
1	&	Random 				& 16.43 	\\
2	&	CR					& 24.31	\\
3	&	CR + vEmbed 			& 34.61	\\
4	&	CR + vAlign			& 19.24	\\
5	&	CR + Look-up	 (LU)	& 29.56 	\\
\hline
& Single Causal Models 		& \\
\hline
6	&	CR + cEmbedBi       & 31.32\\
7	&	CR + cEmbedBiNoise  & 30.15 \\ 
8	&	CR + cAlign  		& 23.49 \\
9	&	CR + cCNN  			& 24.66	\\
\hline
& Model Combinations & \\
\hline
10	&	CR + vEmbed + cEmbedBi		& 37.08$^{*}$	\\ 
11	& 	CR + vEmbed + cEmbedBiNoise 	& 35.50$^*$	\\ 
12	&	CR + vEmbed + cEmbedBi + LU	& 36.75$^{*}$	\\ 
13	&	CR + vEmbed + cAlign		& 	34.31 	\\ 
14	&	CR + vEmbed + cCNN		& 	33.45 \\
\hline
& Model Stacking & \\
\hline
15	& 	CR + vEmbed + cEmbedBi + cEmbedBiNoise 	& {\bf 37.28$^{*}$}	\\ 

\end{tabular}
\end{footnotesize}
\vspace{-1mm}
\caption{{\footnotesize Performance in the QA evaluation, measured by precision-at-one (P@1).  The ``Bi'' suffix indicates a bidirectional model; the ``Noise'' suffix indicates a model that is noise aware. $^*$  indicates that the difference between the corresponding model and the CR + vEmbed baseline is statistically significant ($p < 0.05$), 
determined through a one-tailed bootstrap resampling test with 10,000 iterations. }} 
\label{tab:QA}
\vspace{-8mm}
\end{center}
\end{table}

\subsection{Results}
The overall results are summarized in Table~\ref{tab:QA}.
Lines 1--5 in the table show that each of our baselines performed better than CR by itself, except for vAlign, suggesting that the vanilla alignment model does not generate accurate predictions for causal questions.
The strongest baseline was CR + vEmbed (line 3), the vanilla embeddings trained over Gigaword, at 34.6\% P@1. For this reason, we consider this to be the baseline to ``beat'', and perform statistical significance of all proposed models with respect to it. 

Individually, the cEmbedBi model is the best performing of the causal models.  While the performance of cAlign in the direct evaluation was comparable to that of cEmbedBi, here it performs far worse (line 6 vs 8), suggesting that the robustness of embeddings is helpful in QA.  Notably, despite the strong performance of the cCNN in the low-recall portion of the PR curve in the direct evaluation, here the model performs poorly (line 9).

No individual causal model outperforms the strong vanilla embedding baseline (line 3), likely owing to the reduction in generality inherent to building task-specific QA models.
However, comparing lines 6--9 vs. 10--14 shows that the vanilla and causal models are capturing different and complementary kinds of knowledge (i.e., causality vs. association through distributional similarity), and are able to be combined to increase overall task performance (lines 10--12).  These results highlight that QA is a complex task, where solving methods need to address the many distinct information needs in question sets, including both causal and direct association relations.  This contrasts with the direct evaluation, which focuses strictly on causality, and where the vanilla embedding baseline performs near chance. This observation highlights one weakness of word similarity tasks: their narrow focus may not directly translate to estimating their utility in real-world NLP applications. 

Adding in the lookup baseline (LU) to the best-performing causal model does not improve performance (compare lines 10 and 12), suggesting that the bidirectional causal embeddings subsume the contribution of the LU model.  
cEmbedBi (line 10) also performs better than cEmbedBiNoise (line 11). We conjecture that the ``noise'' filtered out by cEmbedBiNoise contains distributional similarity information, which is useful for the QA task.  cEmbedBi vastly outperforms cCNN (line 14), suggesting that strong overall performance across the precision-recall curve better translates to the QA task.  We hypothesize that the low cCNN performance is caused by insufficient training data, preventing the CNN architecture from generalizing well. 

Our best performing overall model combines both variants of the causal embedding model (cEmbedBi and cEmbedBiNoise), reaching a P@1 of 37.3\%, which shows a 7.7\% relative improvement over the strong CR + vEmbed baseline.

\begin{table}[t!]
\begin{center}
\begin{footnotesize}
\begin{tabular}{ll}
\hline
Error/observation 	& \% Q \\
\hline
Both chosen and gold are equally good answers 	& 	45\% \\ 
Causal max similarity of chosen is higher		&	35\% \\
Vanilla overall similarity of chosen is higher	&	35\% \\
Chosen answer is better than the gold answer		&	25\% \\
The question is very short / lacks content words	&	15\%	 \\
Other 											&	10\% \\
\end{tabular}
\end{footnotesize}

\caption{{\footnotesize Results of an error analysis performed on a random sample of 20 incorrectly answered questions showing the source of the error and the percentage of questions that were affected. Note that questions can belong to multiple categories. }} 
\label{tab:ea}
\vspace{-8mm}
\end{center}
\end{table}

\subsection{Error Analysis}
\label{sec:erroranalysis}

We performed an error analysis to gain more insight into our model as well as the source of the remaining errors.  For simplicity, we used the combination model CR + vEmbed + cEmbedBi. Examining the model's learned feature weights, we found that the vanilla overall similarity feature had the highest weight, followed by the causal overall similarity and causal maximum similarity features.  This indicates that even in causal question answering, the overall \emph{topical} similarity between question and answer is still useful and complementary to the causal similarity features.

To determine sources of error, we randomly selected 20 questions that were incorrectly answered and analyzed them according to the categories shown in Table \ref{tab:ea}.  We found that for 70\% of the questions, the answer chosen by our system was as good as or better than the gold answer, often the case with community question answering datasets.

Additionally, while the maximum causal similarity feature is useful, it can be misleading due to embedding noise, low-frequency words, and even the bag-of-words nature of the model (35\% of the incorrect questions).  For example, in the question \emph{What are the effects of growing up with an older sibling who is better than you at everything?}, the model chose the answer \emph{...You are you and they are them - you will be better and different at other things...}  largely because of the high causal similarity between (\emph{grow} $\rightarrow$ \emph{better}).  While this could arguably be helpful in another context, here it is irrelevant, suggesting that future improvement could come from models that better incorporate textual dependencies.

\vspace{-1mm}
\section{Conclusion}
\vspace{-1mm}

We presented a framework for creating customized embeddings tailored to the information need of causal questions.  We trained three popular models (embedding, alignment, and CNN) using causal tuples extracted with minimal supervision by bootstrapping cause-effect pairs from free text, and evaluated their performance both directly (i.e., the degree to which they capture causality), and indirectly (i.e., their real-world utility on a high-level question answering task). 

We showed that models that incorporate a knowledge of causality perform best for both tasks. 
Our analysis suggests that the models that perform best in the real-world QA task are those that have consistent performance across the precision-recall curve in the direct evaluation.
In QA, where the vocabulary is much larger, precision must be balanced with high-recall, and this is best achieved by our causal embedding model.  Additionally, we showed that vanilla and causal embedding models address different information needs of questions, and can be combined to improve performance. 

Extending this work beyond causality, we hypothesize that additional embedding spaces customized to the different information needs of questions would allow for robust performance over a larger variety of questions, and that these customized embedding models should be evaluated both directly and indirectly to accurately characterize their performance.

\section*{Resources}
All code and resources needed to reproduce this work are  available at \url{http://clulab.cs.arizona.edu/data/emnlp2016-causal/}.

\section*{Acknowledgments}
We thank the Allen Institute for Artificial Intelligence for funding this work.
Additionally, this work was partially funded by the Defense Advanced
Research Projects Agency (DARPA) Big Mechanism
program under ARO contract W911NF-14-1-0395.

\newpage
\bibliography{emnlp2016}
\bibliographystyle{emnlp2016}

\end{document}